\documentclass[twoside,11pt]{article}
\usepackage{tikz,pgm,algorithm,algorithmic,bm,breqn,booktabs,subfigure,pgfplots}

\pgfplotsset{compat=1.17}
\usetikzlibrary{arrows,plotmarks,decorations.markings,trees,shapes}
\usepgfplotslibrary{statistics}
\tikzset{dot/.style = {circle, fill, minimum size=#1,inner sep=0pt, outer sep=0pt, fill, circle},dot/.default = 6pt}
\tikzset{dot2/.style = {circle, fill, color=black!40,minimum size=6pt,inner sep=0pt, outer sep=0pt, fill, circle}}
\tikzstyle{a}=[->,>=stealth,dashed]
\tikzstyle{a2}=[->,>=stealth]
\tikzstyle{a3}=[<->,>=stealth]
\usepackage{hyperref,url}
  \hypersetup{  
    bookmarksnumbered
  }
  \hypersetup{  
    breaklinks=false,
    pdfborderstyle={/S/U/W 0},
    citebordercolor=.235 .702 .443,
    urlbordercolor=.255 .412 .882,
    linkbordercolor=.804 .149 .149,
  }
  \hypersetup{ 
    pdfauthor={},
    pdftitle={},
    pdfkeywords={}
  }
\newcommand{\twoFone}{\ensuremath{{}_2F_1}}

\ShortHeadings{Bounding Counterfactuals under Selection Bias}{Zaffalon et al.}
\begin{document}
\title{Bounding Counterfactuals under Selection Bias}
\author{\Name{Marco Zaffalon} \Email{zaffalon@idsia.ch}\\
\addr IDSIA, Lugano (Switzerland)
\and
\Name{Alessandro Antonucci} \Email{alessandro@idsia.ch}\\
\addr IDSIA, Lugano (Switzerland)
\and
\Name{Rafael Caba\~nas} \Email{rcabanas@ual.es}\\
\addr Department of Mathematics, University of Almer\'{i}a, Almer\'{i}a (Spain)
\and
\Name{David Huber} \Email{david.huber@idsia.ch}\\
\addr IDSIA, Lugano (Switzerland)
\and
\Name{Dario Azzimonti} \Email{dario.azzimonti@idsia.ch}\\
\addr IDSIA, Lugano (Switzerland)}
\maketitle
\begin{abstract}
Causal analysis may be affected by selection bias, which is defined as the systematic exclusion of data from a certain subpopulation. Previous work in this area focused on the derivation of identifiability conditions. We propose instead a first algorithm to address both identifiable and unidentifiable queries. We prove that, in spite of the missingness induced by the selection bias, the likelihood of the available data is unimodal. This enables us to use the causal expectation-maximisation scheme to obtain the values of causal queries in the identifiable case, and to compute bounds otherwise. Experiments demonstrate the approach to be practically viable. Theoretical convergence characterisations are provided.
\end{abstract}

\section{Introduction}\label{sec:intro}
Table \ref{tab:study} reports the results of a drug study where gender is taken into account. The study involves 700 patients, but the outcomes for treated females and untreated males (grey counts in the table) are absent because of an issue in the communication protocol. This is an example of \emph{selection bias}: a subpopulation is systematically missing from the sample and this makes any direct data analysis unreliable. (In what follows, we call these data just `biased' for brevity.)

Such situations have been largely studied in the literature (see, e.g., \citet{winship1992models,zaffalon2009a}), but relatively little work has been devoted to analysing the question from a causal perspective. After the pioneering work of \citet{cooper1995causal}, the problem was clearly cast within the framework of structural causal models by \citet{pearl2012solution}. This eventually led to sound and complete graphical and algorithmic conditions for recovering probabilities from biased data \citep{bareinboim2012controlling,bareinboim2015recovering}. These contributions can be understood as an extension to biased data of Pearl's do-calculus, which reduces causal to observational queries in the \emph{identifiable} cases. Yet, when the focus is on counterfactuals, most inferences are unidentifiable even without selection bias, and the problem of bounding such queries under selection bias is basically unexplored.

The goal of this paper is to fill this gap by providing numerical bounds to unidentifiable counterfactual queries under selection bias. To this end, we start from the recently proposed procedure of \citet{zaffalon2021} designed to compute those bounds by an iterated, so-called, causal EM scheme. That approach exploits the unimodality of the marginal likelihood of the (unbiased) data. Here, we prove that, in spite of the additional missingness induced by the selection of data, the marginal likelihood remains unimodal. This allows us to adopt the causal EM scheme to compute the bounds of unidentifiable queries under selection bias. To the best of our knowledge this is the first technique proposed for such a task. 

The paper is organised as follows. In Sect.~\ref{sec:background} we define the basic notation and the necessary background material. The causal EM scheme is reviewed in Sect.~\ref{sec:emcc}, where an extended characterisation of convergence is provided. Our approach is discussed in Sect.~\ref{sec:s-emcc}. Experiments are reported in Sect.~\ref{sec:experiments} and conclusions in Sect.~\ref{sec:conc}. Proofs are gathered in the appendix.
\begin{table}[htp!]
\centering
{\scriptsize
\begin{tabular}{cccr}
\toprule
Treatment ($X$)&Recovery ($Y$)&Gender ($Z$)&$\#$\\
\midrule
0&0&0&\color{black!30}{2}\\
1&0&0&41\\
0&1&0&\color{black!30}{114}\\
1&1&0&313\\
0&0&1&107\\
1&0&1&\color{black!30}{109}\\
0&1&1&13\\
1&1&1&\color{black!30}{1}\\
\bottomrule
\end{tabular}}
\caption{Data from an observational study involving three Boolean variables \citep{mueller2021causes}. True states correspond to treated ($X$), recovered ($Y$) and female ($Z$).}\label{tab:study}
\end{table}\section{Structural Causal Models}\label{sec:background}
We assume the reader to have some familiarity with probability theory and \emph{structural causal models} (SCMs). Let us focus on models with categorical variables. Variable $V$ takes its values from a finite set $\Omega_V$ with $v$ denoting its generic value. $P(V)$ is a \emph{probability mass function} (PMF) over $V$. Notation $P(V|V'):=\{P(V|v')\}_{v'\in\Omega_{V'}}$ is used instead for a \emph{conditional probability table} (CPTs), i.e., a collection of conditional PMFs over a variable indexed by the states of another one. We define an SCM $M$ by a directed graph $\mathcal{G}$ whose nodes are in a one-to-one correspondence with both its endogenous variables $\bm{X}:=(X_1,\ldots,X_n)$ and the exogenous ones $\bm{U}:=(U_1,\ldots,U_m)$. We focus on semi-Markovian models, i.e., $\mathcal{G}$ is assumed to be acyclic. The exogenous variables correspond to the root nodes of $\mathcal{G}$. A PMF $P(U)$ is specified for each $U\in\bm{U}$. A \emph{structural equation} (SE) $f_X$ is instead provided for each $X \in \bm{X}$; this is a map $f_X:\Omega_{\mathrm{Pa}_X} \to \Omega_X$, where $\mathrm{Pa}_X$ are the parents (i.e., the immediate predecessors) of $X$ according to $\mathcal{G}$. To have all the states of $X$ possibly realised, we only consider SEs corresponding to surjective maps. We might also require a \emph{joint surjectivity}, meaning that any $\bm{x}\in\Omega_{\bm{X}}$ can be realised for at least a $\bm{u}\in\Omega_{\bm{U}}$. SCM $M$ induces the following joint PMF:
\begin{equation}\label{eq:joint}
P(\bm{x},\bm{u})=\prod_{X\in\bm{X}} P(x|\mathrm{pa}_X) \prod_{U\in\bm{U}} P(u)\,,
\end{equation}
for each $\bm{x}\in\Omega_{\bm{X}}$ and $\bm{u}\in\Omega_{\bm{U}}$, where the values of the CPTs associated with each $X$ are degenerate, i.e.,
$P(x|\mathrm{pa}_X)=\delta_{x,f_X(\mathrm{pa}_X)}$. These are \emph{fully specified}, or just \emph{full}, SCMs (FSCMs). If the exogenous PMFs are not provided, we say instead that $M$ is a \emph{partially specified} (or \emph{partial}, PSCMs). In that case, endogenous information is assumed to be available in the form of a dataset $\mathcal{D}$ of observations of $\bm{X}$. Given $X \in \bm{X}$ and $x\in\Omega_X$, an \emph{intervention}, denoted as $\mathrm{do}(X=x)$, describes the action of setting $X$ equal to $x\in\Omega_X$. This corresponds to a \emph{surgery} involving the replacement of $f_X$ with a constant map $X=x$, the consequent removal from $\mathcal{G}$ of the arcs pointing to $X$, and letting $X=x$ be
the evidence.

\begin{figure}[htp!]
\centering
\begin{tikzpicture}[scale=1.0]
\node[dot,label=above left:{$U$}] (ux)  at (1,0) {};
\node[dot,label=above right:{$V$}] (uy)  at (5,0) {};
\node[dot,label=above left:{$W$}] (uz)  at (2,0.7) {};
\node[dot,label=above left:{$X$}] (x)  at (2,0) {};
\node[dot,label=above right:{$Y$}] (y)  at (4,0) {};
\node[dot,label=above left:{$Z$}] (z)  at (3,0.7) {};
\draw[a] (ux) -- (x);
\draw[a] (uy) -- (y);
\draw[a] (uz) -- (z);
\draw[a2] (x) -- (y);
\draw[a2] (z) -- (x);
\draw[a2] (z) -- (y);
\end{tikzpicture}
\caption{A Markovian SCM.}\label{fig:scm}
\end{figure}
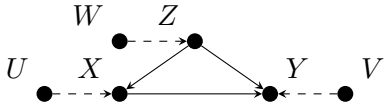

Fig.~\ref{fig:scm} represents the graph of the SCM $M$ for the study with outcomes in Tab.~\ref{tab:study}. The model is \emph{Markovian}, i.e., each endogenous variable has a single exogenous parent. We set \emph{conservative} SEs \citep{zaffalon2020}, i.e., the exogenous states are indexing all the possible (deterministic) functions from the endogenous parents of an endogenous variable to such a variable (see also \citet{zhang2021counterfactual} for an extension of this concept). Accordingly, we have $|\Omega_U|=4$, $|\Omega_W|=2$, $|\Omega_V|=16$.

Causal queries on SCMs involve endogenous variables. Following Pearl's causal hierarchy, we distinguish three kinds of queries. Observational queries, such as $P(X=1|Y=1)$, involving the evaluation of marginal or posterior queries for some endogenous variables given an observation of some other ones; interventional queries such as $P(Y=1|\mathrm{do}(X=1))$, where the causal effect of an intervention is considered; and counterfactual queries where the same variable is intervened in a state and observed (or intervened) in another one. Classical counterfactual queries for models with Boolean endogenous variables are the probability of necessity $\mathrm{PN}:=P(Y_{X=0}=0|X=1,Y=1)$, sufficiency $\mathrm{PS}:=P(Y_{X=1}=1|X=0,Y=0)$ and necessity and sufficiency $\mathrm{PNS}:=P(Y_{X=0}=0,Y_{X=1}=1)$. Subscripts are used to denote interventions related to counterfactuals and, in those queries, $\mathcal{G}$ is so that $X$ topologically precedes $Y$. E.g., in medical applications PNS can describe the fraction of the population that recovers if cured and does not otherwise. Observational queries can be computed  by standard Bayesian network inference tools. In FSCMs, interventional queries can similarly be addressed after standard surgery, while in PSCMs, techniques such as do-calculus allow for a reduction to observational queries in specific cases. Counterfactual queries in FSCMs typically require the application of Bayesian network inference to auxiliary structures like twin networks or counterfactual graphs \citep{shpitser2012counterfactuals}. The problem of addressing counterfactual queries in PSCMs is discussed in the next section.\section{Addressing Partial Identifiability (without Selection Bias)}\label{sec:emcc}
As we have remarked already, the general problem of computing counterfactuals can be reduced to standard Bayesian network inference in FSCMs. The same cannot be done with PSCMs: tools like do-calculus can reduce the causal query to an observational one only in the \emph{identifiable} case. For unidentifiable queries, one needs to compute probability bounds. 

As an example consider the PNS for the SCM in Fig.~\ref{fig:scm} discussed in Sect.~\ref{sec:background} with the endogenous data in Tab.~\ref{tab:study}. This is an unidentifiable query, whose bounds $0 \leq \mathrm{PNS} \leq 0.01458$ have been obtained by \citet{mueller2021causes} with some ad-hoc derivation; an exact characterisation of these bounds has been provided by \citet{zaffalon2020}. 

More generally speaking, various approaches have recently been proposed to approximate unidentifiable bounds (see, e.g., \citet{zhang2021counterfactual,zaffalon2021}). Yet, there appears to be no viable technique to compute general counterfactual inference under selection bias. This will be the topic of the next section, which will show how EMCC has a natural extension to problems of selection bias. Before that, we recall the basics of EMCC (Sect.~\ref{sec:emcc_alg}) and derive improved credible intervals for it (Sect.~\ref{sec:confidence}).

\subsection{EMCC \citep{zaffalon2021}}\label{sec:emcc_alg}
Given a PSCM and a dataset of (unbiased) endogenous observations,
Alg.~\ref{alg:cem} yields an FSCM compatible with the PSCM (provided that there is one): such an FSCM is obtained by adding PMFs to the exogenous variables of the PSCM in such a way that the resulting model can generate the distribution of the available data.

In practice, given an initialisation $\{P_0(U)\}_{U\in\bm{U}}$, the EM algorithm consists in regarding the posterior probability $P_0(u|\bm{x})$ as a \emph{pseudo-count} for $(u,\bm{x})$, for each $\bm{x}\in\mathcal{D}$,  $u\in\Omega_U$ and $U\in\bm{U}$ (E-step). A new estimate $P_1(u):=\sum_{\bm{x}\in\mathcal{D}} \frac{P_0(u|\bm{x})}{|\mathcal{D}|}$ is consequently obtained (M-step). The scheme is iterated until convergence. 
Subroutine ${\tt initialisation}$ (line 1) provides a random initialisation of the exogenous PMFs, $\epsilon$ is the threshold to evaluate parameter convergence w.r.t. a metric $\delta$ (line 3). The FSCM returned by Alg.~\ref{alg:cem} can indeed be proved compatible with the PSCM. 

Computing a causal query in this FSCM gives the exact value of the query in such an identifiable case, which then corresponds to a point of the interval for the unidentifiable case represented by the PSCM \citep{zaffalon2021}. An inner approximation of such an interval is eventually achieved by iterating the EM scheme above with variable initialisations. This corresponds to sampling the space of FSCMs compatible with the given PSCM.

\begin{algorithm}[htp!]
\caption{EMCC: given SCM $M$ and dataset $\mathcal{D}$ returns $\{P(U)\}_{U\in\bm{U}}$.}
\begin{algorithmic}[1]
\STATE $\{P_0(U)\}_{U\in\bm{U}} \leftarrow {\tt initialisation}(M)$
\STATE $t \leftarrow 0$
\WHILE{$\delta[\{P_{t+1}(U)\}_{U\in\bm{U}},\{P_t(U)\}_{U\in\bm{U}}]> \epsilon$}
\FOR{$U\in \bm{U}$}
\STATE $P_{t+1}(U) \leftarrow |\mathcal{D}|^{-1}\sum_{\bm{x} \in \mathcal{D}} P_t(U|\bm{x})$
\STATE $t \leftarrow t+1$
\ENDFOR
\ENDWHILE
\end{algorithmic}\label{alg:cem}
\end{algorithm}

\subsection{A New Credible Interval for Unidentifiable Inference}\label{sec:confidence}
A characterisation of the EMCC accuracy in terms of credible intervals has been provided by \citet{zaffalon2021}.  Let $\rho:=\{\pi_i\}_{i=1}^k$ denote the output of $k$ EMCC iterations, then $[a,b]$ is the interval induced by $\rho$, i.e., $a:=\min_{i=1}^k \pi_i$ and
$b:=\max_{i=1}^k \pi_i$. If $[a^*,b^*]$ are the exact probability bounds for a causal query,
by construction, we have $a^*\leq a \leq b \leq b^*$. 
The following equality holds:
\begin{equation}\label{eq:bounds_uniform}
P\left(
a-\varepsilon L\leq
a^*\leq b^* \leq b+\varepsilon L\,\bigg|\, \rho \right) =
\frac{
1+(1+2\varepsilon)^{2-k}-2(1+\varepsilon)^{2-k}}{
(1-L^{k-2})
-(k-2)(1-L)L^{k-2}}
\,,
\end{equation}
where $L:=(b-a)$ and $\varepsilon:=\delta/(2L)$ is the relative error at each extreme of the interval obtained as a function of the absolute allowed error $\delta\in(0,L)$. Eq.~\eqref{eq:bounds_uniform} is derived under a uniformity assumption for the distributions of the points in $\rho$. Here we obtain a new result based on weaker assumptions.
\begin{theorem}\label{th:newbounds}
Assume that the EMCC runs in $\rho$ are distributed as a four parameter Beta distribution, i.e., $\pi_i \sim Beta(\alpha,\beta,a^*, b^*)$, for each $i=1, \ldots, k$. The following equality holds:
\begin{equation}\label{eq:confidenceBeta}
P\left(
a-\varepsilon L\leq
a^*\leq b^* \leq b+\varepsilon L\,\bigg|\, \rho \right) 
=  
\dfrac{\displaystyle\int_{0}^{\delta/2}\int_{0}^{\delta/2} P(x,y; L,\alpha, \beta,k) \, \mathrm{d}x \, \mathrm{d}y}{\displaystyle\int_{0}^{a+(1-b)}\int_{0}^{a+(1-b)-y}P(x,y; L,\alpha, \beta,k) \, \mathrm{d}x \, \mathrm{d}y}\,,
\end{equation}
where $P(x,y; L,\alpha, \beta, k)$ is equal to
\begin{equation}\label{eq:newbounds}
\left(\dfrac{(L+x)^\alpha \twoFone(\alpha,1-\beta,\alpha+1,\frac{L+x}{L+x+y})- x^\alpha \twoFone(\alpha,1-\beta,\alpha+1,\frac{x}{L+x+y})}{\alpha (L+x+y)^{\alpha} B(\alpha,\beta)}\right)^k\,,
\end{equation}
$\twoFone$ is the Gaussian (ordinary) hypergeometric function and $B(\alpha,\beta)$ is the beta function evaluated at $\alpha,\beta>0$. 
\end{theorem}
The above theorem is a proper extension of the original EMCC characterisation, as proved by the following:
\begin{corollary}\label{cor:newbounds}
If the EMCC runs in $\rho$ are uniformly distributed in $[a^*, b^*]$,
i.e., for each $i=1, \ldots, k$, $\pi_i \sim Beta(1,1,a^*, b^*)$, then we recover Eq.~\eqref{eq:bounds_uniform}.
\end{corollary}
After any $k$ EMCC runs, we can compute $a$ and $b$ and, given a relative error $\delta$ we regard as acceptable, obtain the probability in Eq.~\eqref{eq:newbounds} by estimating the values of the parameters $\alpha$ and $\beta$ with a maximum likelihood procedure over the $k$ values collected in $\rho$. If the corresponding probability is sufficiently high, we stop iterating EMCC, otherwise we keep iterating the procedure to collect new points for $\rho$ and achieve greater probabilities. 

A specialisation of the bound in Eq.~\eqref{eq:bounds_uniform} can easily be obtained to decide if the query is identifiable (i.e., $a^*=b^*$), when we obtain the same value of $\pi_i$ for all $i=1, \ldots, k$. 
\begin{corollary}\label{cor:boundaequalb}
If $a=b$, i.e., all $k$ runs in $\rho$ are equal, then
$P(a^*=b^*|\rho) =1+ 9/3^k-8/2^k$.
\end{corollary}
This implies that nine equal runs make identifiability probable with 99\% confidence.\section{Addressing Partial Identifiability under Selection Bias}\label{sec:s-emcc}
To model the effect of selection bias, we define a \emph{selector} $S$, i.e., a Boolean variable that is true for selected states of $\bf{X}$ and false otherwise. We consider deterministic selectors, i.e., $S:=g(\bm{X})$ with $g:\Omega_{\bm{X}}\to \{0,1\}$. E.g., the bias discussed in Sect.~\ref{sec:intro} for the data in Tab.~\ref{tab:study} corresponds to selector $S:= Z \vee \neg X$. $S$ can be embedded in SCM $M$ as a common child of the related endogenous variables, with $g$ being the SE of $S$, thus acting as an additional endogenous variable with endogenous parents only. 

Given a dataset $\mathcal{D}$ of observations of $\bm{X}$, $S$ partitions $\mathcal{D}$ in: $\mathcal{D}_0$ where $S=0$ and the endogenous observations are missing; and $\mathcal{D}_1$ where $S=1$ and the observations remain as in $\mathcal{D}$. We use notation $\mathcal{D}_S := \mathcal{D}_0 \cup \mathcal{D}_1$ and, for cardinalities, $d_0:=|\mathcal{D}_0|$, $d_1:=|\mathcal{D}_1|$, with $d_0+d_1=d:=|\mathcal{D}|=|\mathcal{D}_S|$. Note that $\mathcal{D}_0$ contains $d_0$ identical records. An example is in Fig.~\ref{fig:sb}.
\begin{figure}[htp!]
\centering
\begin{tikzpicture}
\node (tab) at (4,1) {\scriptsize
\begin{tabular}{ccccc}
\toprule
$Z$&$X$&$Y$&$S$&\#\\
\midrule
0&0&0&0&2\\
0&0&1&0&114\\
0&1&0&1&41\\
0&1&1&1&313\\
1&0&0&1&107\\
1&0&1&1&13\\
1&1&0&0&109\\
1&1&1&0&1\\
\bottomrule
\end{tabular}
};
\node (tab1) at (11,1.3) {\scriptsize
\begin{tabular}{ccc ccc cc}
\hline 
$W$&$U$&$V$ &$Z$&$X$&$Y$ &$S$&\#\\
\hline 
*&*&*& 0&1&0& 1&41\\
*&*&*& 0&1&1& 1&313\\
*&*&*& 1&0&0& 1&107\\
*&*&*& 1&0&1& 1&13\\
\hline 
\end{tabular}
};
\node (tab0) at (11,-0.2) {\scriptsize
\begin{tabular}{cccccccc}
\hline 
$W$&$U$&$V$ &$Z$&$X$&$Y$ &$S$&\#\\
\hline 
*&*&*& *&*&*& 0&226\\
\hline 
\end{tabular}
};
\node (d) at (1.6,1) {\Large $\mathcal{D}$};
\node (d1) at (14.8,1.2) {\Large $\mathcal{D}_1$};
\node (d0) at (14.8,0.0) {\Large $\mathcal{D}_0$};
\draw[a2] (tab) -- (tab1);
\draw[a2] (tab) -- (tab0);
\end{tikzpicture}
\caption{Selecting the data in Tab.~\ref{tab:study} with $S:=Z\vee \neg X$.}\label{fig:sb}
\end{figure}

The marginal log-likelihood $LL$ of $\mathcal{D}_S$ according to $M$ is:
\begin{equation}\label{eq:llx}
LL:=\log P(\mathcal{D}_S)= \log P(\mathcal{D}_0) + \log P(\mathcal{D}_1) = 
d_0 \log {P}(S=0) + \sum_{\bm{x}\in\mathcal{D}_1} \log {P}(\bm{x})\,.
\end{equation}
As Eq.~\eqref{eq:llx} is not the likelihood of a dataset of complete endogenous observations for an SCM,  the characterisation provided by \cite{zaffalon2021} to justify the application of EMCC does not directly hold. 

Yet, Eq.~\eqref{eq:llx} contains the sum of two log-likelihoods for disjoint datasets. This leads us to define a simple upper bound $LL^*$ for $LL$ by using maximum-likelihood estimates separately for $\mathcal{D}_0$ and $\mathcal{D}_1$, i.e.,
\begin{equation}\label{eq:ll_star}
LL^*:= d_0 \log \hat{P}(S=0) + \sum_{\bm{x}\in\mathcal{D}_1} \log \hat{P}(\bm{x})\,,
\end{equation}
where $\hat{P}(S=0):=d_0/d$ and, for each $\bm{x}\in\mathcal{D}_1$, $\hat{P}(\bm{x})$ is obtained from the counts in $\mathcal{D}_1$ by means of the endogenous factorisation induced by the \emph{c-components} of $M$ as in Eqs.~(2) and~(6) of \citet{zaffalon2021}.
As an explicit function of the exogenous PMFs, the log-likelihood of $\mathcal{D}_S$ is instead:
\begin{equation}\label{eq:llu}
LL[\{P(U)\}_{U\in\bm{U}}] :=
d_0 \log \sum_{\bm{x} \in \Omega_{\bm{X}}^{S=0}} \sum_{\bm{u} \in \Omega_{\bm{U}}}  \delta_{f(\bm{u}),\bm{x}} P(\bm{u})
+
\sum_{\bm{x} \in \mathcal{D}_1} \log \sum_{\bm{u} \in \Omega_{\bm{U}}} \delta_{f(\bm{u}),\bm{x}} P(\bm{u})
\,,
\end{equation}
where $\Omega_{\bm{X}}^{(S=0)}:=g^{-1}(S=0)$ and 
$f:\Omega_{\bm{U}}\to\Omega_{\bm{X}}$ is the \emph{joint} SE we build from $\{f_X\}_{X\in\bm{X}}$.
Remember that we should only be after FSCMs that are compatible with the given PSCM (also called \emph{$M$-compatible}), as mentioned in Sect.~\ref{sec:emcc_alg}. In the present case, $M$-compatibility corresponds to imposing the following constraints:
\begin{equation}\label{eq:sm_compat1}
\sum_{\bm{x} \in \Omega_{\bm{X}}^{S=0}} \sum_{\bm{u} \in \Omega_{\bm{U}}} \delta_{f(\bm{u}),\bm{x}} P(\bm{u}) = \hat{P}(S=0)\,,
\end{equation}
and, for each $\bm{x} \in \mathcal{D}_1$,
\begin{equation}\label{eq:sm_compat2}
\sum_{\bm{u} \in \Omega_{\bm{U}}} \delta_{f(\bm{u}),\bm{x}}  P(\bm{u})  = \hat{P}(\bm{x})\,.
\end{equation}
These compatibility conditions should be self-explanatory in that the models we look for should be capable of generating the distribution of the data, besides being an extension of the given PSCM. It can anyway be shown that they are just the form that the notion of $M$-compatibility, given in \cite{zaffalon2021}, takes when it is applied to the present context (we omit the proof).

Checking such a compatibility is equivalent to a simple evaluation of the function in Eq.~\eqref{eq:llu} as stated by the following result.
\begin{theorem}\label{th:unimodal}
As a function of $\{P(U)\}_{U\in\bm{U}}$, the log-likelihood in Eq.~\eqref{eq:llu} has no local maxima and a global maximum equal to the value $LL^*$ in Eq.~\eqref{eq:ll_star}. Such a maximum is achieved if and only if the $M$-compatibility constraints in Eqs.~\eqref{eq:sm_compat1} and~\eqref{eq:sm_compat2} are satisfied.
\end{theorem}
As a consequence of Thm.~\ref{th:unimodal}, the application of Alg.~\ref{alg:cem}
to PSCM $M$ and data $\mathcal{D}_S$, returns only FSCMs whose log-likelihood takes its global maximum $LL^*$ because of the EM properties and hence satisfy the compatibility constraints. Causal queries computed with those FSCMs are consequently inner points of the exact interval for the partially identifiable query. The more points the better the approximation. 

Note that as the data in $\mathcal{D}_0$ are $d_0$ instances of the same observation $S=0$, when coping with selection bias, line 5 of Alg.~\ref{alg:cem} rewrites as:
\begin{equation}\label{eq:newline}
P_{t+1} \leftarrow \dfrac{d_0 P_t(U|S=0)+\sum_{\bm{x} \in \mathcal{D}_1} P_t(U|\bm{x})}{
(d_0+d_1)}\,.
\end{equation}

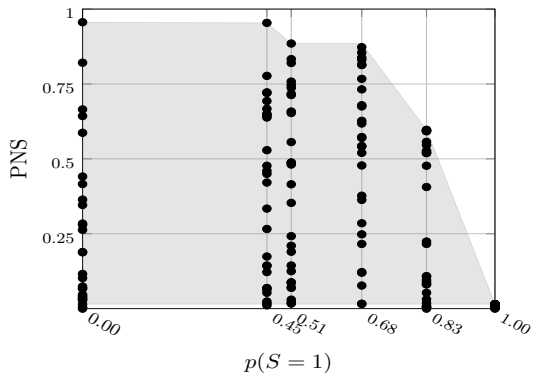
\begin{figure}[htp!]
\centering
\begin{tikzpicture}[yscale=0.7,xscale=0.8]
\pgfplotsset{every x tick label/.append style={font=\scriptsize, yshift=-0.5ex}}
\pgfplotsset{every y tick label/.append style={font=\scriptsize, yshift=-0.5ex}}
\begin{axis}[xmin = 0, xmax = 1,
ymin = 0, ymax = 1,
xlabel = {$p(S=1)$},
ylabel = {PNS},
x tick label style={rotate=330,anchor=west},
xtick = {1.000,0.834,0.677,0.506,0.447,0.000},
ytick = {0.250,0.500,0.750,1.000},
xticklabels = {1.00,0.83,0.68,0.51,0.45,{\footnotesize 0.00}},
grid = both,
major grid style = {lightgray},
minor grid style = {lightgray!25}]
\addplot[gray,fill,opacity=0.2,sharp plot] table [x ={x}, y = {upper}] {\table}|- (current plot begin);
\addplot[black, only marks] table [x ={x1}, y = {y1}]{\ttable};
\addplot[black, only marks] table [x ={x2}, y = {y2}]{\ttable};
\addplot[black, only marks] table [x ={x3}, y = {y3}]{\ttable};
\addplot[black, only marks] table [x ={x4}, y = {y4}]{\ttable};
\addplot[black, only marks] table [x ={x5}, y = {y5}]{\ttable};
\addplot[black, only marks] table [x ={x6}, y = {y6}]{\ttable};
\end{axis}
\end{tikzpicture}
\caption{PNS bounds for different selection levels (x-axis). Exact bounds (grey) induced by $r=30$ EMCC runs and points associated with each run (black) are depicted.}
\label{fig:pearl_sb}
\end{figure}

As a demonstrative example, Fig.~\ref{fig:pearl_sb} displays the outputs of EMCC for the SCM $M$ described in Sect.~\ref{sec:background} in the presence of a selection bias on the dataset in Tab.~\ref{tab:study}. Different selection mechanisms, including the one discussed in Fig.~\ref{fig:sb} that corresponds to $P(S=1)=0.68$, are considered. The size of the PNS interval for the unselected case $P(S=1)=1$, i.e., $[0,0.1458]$, rapidly increases if fewer data are selected. We also observe that: (i) with no data at all, the PNS interval is very large $[0,0.9564]$ but not vacuous (this means that SEs alone provide some information about the interval); (ii) if records are removed incrementally, the bounds increase in a \emph{monotone} way (these empirical facts have been also observed in all the experiments considered in Sect.~\ref{sec:experiments}); (iii) the intervals are quite large as a consequence of the simple graphical structure considered.

\pgfplotstableread{
x   upper
1   0.015
0.834 0.597
0.677 0.885
0.505 0.885
0.447 0.954
0.00   0.956
}{\table}
\pgfplotstableread{
x1      y1      x2      y2      x3      y3      x4      y4      x5      y5      x6      y6
1.000   0.002   0.834   0.007   0.677   0.016   0.506   0.070   0.447   0.065   0.000   0.000
1.000   0.006   0.834   0.006   0.677   0.077   0.506   0.758   0.447   0.638   0.000   0.441
1.000   0.003   0.834   0.108   0.677   0.216   0.506   0.017   0.447   0.018   0.000   0.047
1.000   0.014   0.834   0.224   0.677   0.873   0.506   0.820   0.447   0.650   0.000   0.005
1.000   0.003   0.834   0.007   0.677   0.679   0.506   0.653   0.447   0.643   0.000   0.821
1.000   0.014   0.834   0.108   0.677   0.570   0.506   0.088   0.447   0.069   0.000   0.034
1.000   0.014   0.834   0.557   0.677   0.767   0.506   0.125   0.447   0.122   0.000   0.038
1.000   0.015   0.834   0.089   0.677   0.732   0.506   0.087   0.447   0.053   0.000   0.029
1.000   0.005   0.834   0.546   0.677   0.520   0.506   0.415   0.447   0.421   0.000   0.043
1.000   0.012   0.834   0.082   0.677   0.838   0.506   0.713   0.447   0.722   0.000   0.416
1.000   0.009   0.834   0.519   0.677   0.675   0.506   0.242   0.447   0.266   0.000   0.280
1.000   0.015   0.834   0.597   0.677   0.627   0.506   0.658   0.447   0.667   0.000   0.263
1.000   0.003   0.834   0.095   0.677   0.248   0.506   0.190   0.447   0.144   0.000   0.116
1.000   0.015   0.834   0.406   0.677   0.815   0.506   0.746   0.447   0.693   0.000   0.665
1.000   0.015   0.834   0.593   0.677   0.573   0.506   0.737   0.447   0.648   0.000   0.643
1.000   0.015   0.834   0.592   0.677   0.543   0.506   0.070   0.447   0.067   0.000   0.188
1.000   0.012   0.834   0.081   0.677   0.122   0.506   0.481   0.447   0.477   0.000   0.189
1.000   0.006   0.834   0.018   0.677   0.377   0.506   0.717   0.447   0.777   0.000   0.587
1.000   0.015   0.834   0.594   0.677   0.478   0.506   0.488   0.447   0.450   0.000   0.346
1.000   0.006   0.834   0.477   0.677   0.285   0.506   0.556   0.447   0.529   0.000   0.067
1.000   0.002   0.834   0.053   0.677   0.120   0.506   0.144   0.447   0.143   0.000   0.102
1.000   0.015   0.834   0.216   0.677   0.855   0.506   0.833   0.447   0.720   0.000   0.364
1.000   0.014   0.834   0.520   0.677   0.541   0.506   0.020   0.447   0.024   0.000   0.068
1.000   0.015   0.834   0.523   0.677   0.813   0.506   0.885   0.447   0.954   0.000   0.956
1.000   0.014   0.834   0.527   0.677   0.618   0.506   0.353   0.447   0.334   0.000   0.030
1.000   0.000   0.834   0.000   0.677   0.016   0.506   0.030   0.447   0.011   0.000   0.014
1.000   0.014   0.834   0.596   0.677   0.831   0.506   0.210   0.447   0.174   0.000   0.074
1.000   0.013   0.834   0.031   0.677   0.363   0.506   0.487   0.447   0.460   0.000   0.284
}{\ttable}

Note that in this limit, we are basically ignoring the data in $\mathcal{D}_1$. If this is the case, the log-likelihood is proportional to $d_0$ and this implies that EMCC gives the same results irrespectively of the size of the (completely missing) dataset.
Let us remark, with reference to Eq.~\eqref{eq:newline}, that the execution of EMCC under selection bias requires the cardinality $d_0$ of $\mathcal{D}_0$. Yet, as these data are unavailable by definition, the value of $d_0$ might be unavailable too. If this is the case, it is in principle possible to estimate $d_0$ from $P(S=0)$ as $d_0/d_1 \simeq P(S=0)/P(S=1)$. If also $P(S=0)$ is not available, a conservative approach may consist in using an upper bound for $P(S=0)$. If this is not possible either, one can still use the (very) conservative approach of taking the limit $P(S=0)\to 1$.

Finally, notice that as in principle $S$ might be a common child of all the variables in $\bm{X}$, implementing SE $g$ as a CPT might lead to an exponential blow-up, this making the inference of $P(U|S=0)$ required by EMCC intractable. In practice, as in \citet{bareinboim2012controlling}, or in our example, the selector might only depend on a subset of $\bm{X}$ of bounded cardinality. This limit might be bypassed by the circuital approach of \citet{darwiche2022causal}.\section{Experiments}\label{sec:experiments}

\paragraph{Benchmark Generation} We assume the endogenous variables of our synthetic SCMs to be Boolean. We randomly generate a directed acyclic graph over $\bm{X}$ with maximum indegree smaller than or equal to $n_p$. This graph is augmented with an exogenous parent for each endogenous node to obtain a Markovian model. The model might become non-Markovian by merging together $q$ (distinct) random pairs of exogenous nodes. In these models we specify \emph{conservative} SEs  \citep[Sect.~2]{zaffalon2021}, thus obtaining a PSCM for each graph. For each $U\in \bm{U}$, we sample a PMF $P(U)$ to obtain an FSCM $M'$. Model $M'$ is used to sample $d$ complete observations of both $\bm{X}$ and $\bm{U}$. We denote by $\mathcal{D}$ the corresponding dataset of endogenous observations. Only $M$-compatible datasets are considered \citep[Sect.~5]{zaffalon2021}. An iterative procedure is applied to the exogenous states that are removed until it is possible to do that while preserving such a compatibility. Given dataset $\mathcal{D}$ and the PSCM $M$, we consider three endogenous variables $X,Y,Z$ so that $X$ is a root node in $\mathcal{G}$, $Y$ is a leaf and $Z$ is an internal node. $P(S=1|X,Y,Z)$ is set to one on some of the eight states of $(X,Y,Z)$ and zero elsewhere. We estimate $P(S=1)$ from the instances of $\mathcal{D}$ that are selected by that particular mechanism. Our benchmark includes 115 SCMs generated by the above procedure with $n_p=3$, $|\bm{X}|\in\{4,5,6,7,8\}$ and $|\mathcal{D}|=1000$ and, for non-Markovian models, $q\in\{2,\ldots,|\bm{X}|/2\}$. 

\begin{figure}[ht]
\centering
\subfigure[]{\begin{tikzpicture}[scale=0.5]
\begin{axis}[
boxplot/draw direction=y,
legend plot pos=right,
legend entries = {Markovian, Non-Markovian},
legend to name={legend},
name=border,
ylabel={$\mathrm{RRMSE}_r$},
ymajorgrids=true,
every axis plot/.append style={fill,fill opacity=0.5},
boxplot={draw position={1/3 + floor(\plotnumofactualtype/2) + 1/3*mod(\plotnumofactualtype,2)},box extend=0.21},
x=3cm,
xtick={0,1,2,...,50},
x tick label as interval,
xticklabels={{$r$ = 10},{$r$ = 20},{$r$ = 30}},
x tick label style={text width=8cm,align=center},
cycle list={{black!50},{black}}]
\addplot+ table[row sep=\\,y index=0] {data 0.1652587703846768 \\ 0.2631653869927064 \\ 0.06489200777989865 \\ 0.560575455811918 \\ 0.0\\}; 
\addplot+ table[row sep=\\,y index=0] {data 0.09391058987900334 \\ 0.1809333984848809 \\ 0.03899377435135172 \\ 0.39384283468517467 \\ 0.0\\}; 
\addplot+ table[row sep=\\,y index=0] {data
0.0875459996812838 \\ 0.17901051722934244 \\ 0.030266664539692248 \\ 0.4021262962638177 \\ 0.0\\}; 
\addplot+ table[row sep=\\,y index=0] {data\\
0.04376223459897535 \\ 0.11041738000477974 \\ 0.0069363867244947 \\ 0.2656388699252073 \\ 0.0\\}; 
\addplot+ table[row sep=\\,y index=0] {data\\
0.0521859143716762 \\ 0.1205262426740707 \\ 0.00746120929114015 \\ 0.2901237927484665 \\ 0.0\\}; 
\addplot+ table[row sep=\\,y index=0] {data\\
0.0167682819539628 \\ 0.06309879220042997 \\ 0.0015062337401717501 \\ 0.15548762989081727 \\ 0.0\\}; 
\node[] at (2.4,0.5) {\ref{legend}};
\end{axis}
\end{tikzpicture}\label{fig:acc}}
\subfigure[]{
\begin{tikzpicture}[scale=0.5]
\begin{axis}[
boxplot/draw direction=y,
name=border,
ylabel={$\mathrm{RRMSE}_r'$},
boxplot={draw position={1/3 + floor(\plotnumofactualtype/3) + 1/3*mod(\plotnumofactualtype,3)},box extend=0.13},
every axis plot/.append style={fill,fill opacity=0.2},
x tick label style={align=center,rotate=-20},
xtick={1/3,2/3,3/3,4/3,5/3,6/3},
xticklabels={0,0-0.2,0.2-0.4,0.4-0.6,0.6-0.8,0.8-1.0},
cycle list={{black},{black},{black},{black},{black},{black}}]
\addplot+ table[row sep=\\,y index=0]
{data\\0.559\\0.649\\0.454\\0.941\\0.163\\};
\addplot+ table[row sep=\\,y index=0] {data\\0.523\\0.618\\0.411\\0.928\\0.101\\};
\addplot+ table[row sep=\\,y index=0] {data\\0.466\\0.592\\0.294\\0.931\\0.025\\};
\addplot+ table[row sep=\\,y index=0] {data\\0.431\\0.503\\0.179\\0.95\\0.000\\};
\addplot+ table[row sep=\\,y index=0] {data\\0.313\\0.485\\0.099\\1.0\\0.000\\};
\addplot+ table[row sep=\\,y index=0] {data\\0.123\\0.440\\0.051\\1.0\\0.000\\};
\end{axis}
\end{tikzpicture}\label{fig:gap}}
\caption{(a) EMCC accuracy for different numbers of runs. (b) Difference between the bounds of biased and unbiased data w.r.t. $P(S=1)$ (x-axis).}\label{fig:emcc_acc}
\end{figure}
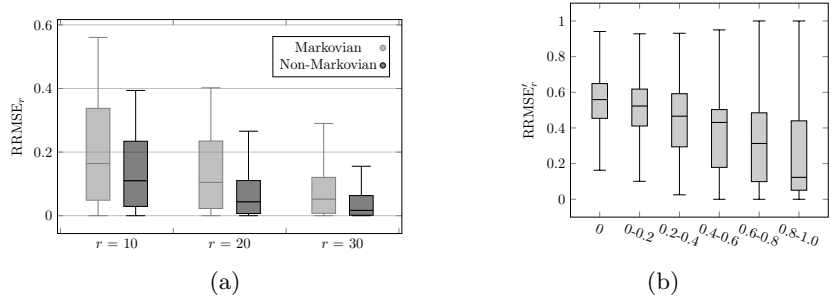

\paragraph{Results} The proposed adaptation of EMCC to cope with selection bias has been implemented within the CREDICI library \citep{credici} and the code to reproduce models and experiments is freely available.\footnote{\url{https://github.com/IDSIA-papers/2022-PGM-selection}.} Given SCM $M$, the biased dataset $\mathcal{D}_1$, and the number $d_0$ of unavailable records, we compute the PNSs $\{\pi_i\}_{i=1}^r$ in the twin networks induced by the FSCMs returned by $r$ EMCC runs. This induces interval $[a_r,b_r]$, where  $a_r:=\min_{i=1}^r \pi_i$ and $b_r:=\max_{i=1}^r \pi_i$. The ground-truth interval $[a^*,b^*]$ for the same setup is obtained as the limit of a large (namely $r_{\mathrm{max}}=80$) number or runs (cf. Thm.~\ref{th:newbounds}). For the unbiased case, we use the method in \cite{zaffalon2020} to evaluate the interval $[a^*,b^*]$. The quality of the (inner) approximation of $[a^*,b^*]$ provided by $[a_r,b_r]$ can be described by the root mean square error of the difference between the upper bounds and that involving the lower bounds. We normalised that by the size $b^*-a^*$ of the true interval in order to properly compare results for intervals of different sizes. Such a \emph{relative} RMSE, i.e.,
\begin{equation}\label{eq:rrmse}
\mathrm{RRMSE}_r := \sqrt{\frac{(a_r-a^*)^2+(b_r-b^*)^2}{2(b^*-a^*)^2}}\,,
\end{equation}
is computed for each experiment. Boxplots are finally used to visualise the aggregated results. Fig.~\ref{fig:acc} depicts such errors for increasing numbers of runs, separately for Markovian (360 experiments) and non-Markovian (306 experiments) models. As expected, more EMCC runs give smaller errors and this happens both for Markovian and non-Markovian models. With few runs ($r=30$) we obtain mean relative errors smaller than 8\% for Markovian models and 5\% for non-Markovian ones in an average time of $7.6$ seconds per run. Such a good accuracy is consistent with the results in Sect.~\ref{sec:confidence}. The difference between Markovian and non-Markovian models appears to be due to the fact that $n_{\bm{U}}:=\prod_{U\in\bm{U}} |\Omega_U|$ is smaller for non-Markovian cases. Those differences are not detected when comparing models w.r.t. different treewidth or interval sizes. In Fig.~\ref{fig:gap} we compare the EMCC intervals ($r=30$) with the intervals obtained with the corresponding unbiased data. The same metric as in Eq.~\eqref{eq:rrmse} is considered, with the interval under selection bias playing the role of the ground truth, and results grouped by different ranges for $P(S=1)$. As expected higher selection-bias levels make the intervals larger, and this behaviour seems to be monotone.

\section{Conclusions}\label{sec:conc}
We have proposed a first algorithm to numerically bound counterfactual queries under selection bias. The algorithm is approximate, being based on the EM, and yields consistently accurate results for a relatively low number of EM runs. As future work, we would like to extend the support to non-deterministic selection mechanisms, while using tractable circuits for the modelling of the deterministic case.

\appendix

\section*{Proofs}
\begin{proof}[of Thm.~\ref{th:newbounds}]
Consider the l.h.s. of Eq.~\eqref{eq:confidenceBeta}. The corresponding joint density is:
\begin{equation}
P\left( \Delta_a \leq \dfrac{\delta}{2}, \Delta_b \leq \dfrac{\delta}{2}, \rho \right) = \gamma \int_{0}^{\frac{\delta}{2}}\int_{0}^{\frac{\delta}{2}} P(\rho | \Delta_a = x, \Delta_b = y) \mathrm{d}x \mathrm{d}y\,,
\end{equation}
where a uniform prior is considered
and $\Delta_a := (a-a^*)$ and $\Delta_b := (b^*-b)$. As $\pi_i \sim \operatorname{Beta}(\alpha, \beta, a^*, b^*)$,
$P(\pi_i \in [a,b]|\Delta_a=x,\Delta_b=y)$ is $P(\pi_i \in [a,b]|a^*=a-x,b^*=b+y)$ and: 
\begin{equation}\footnotesize
\int_{a}^{b} \dfrac{(\rho-a+x)^{\alpha-1}(b+y-\rho)^{\beta-1}}{(b-a+y+x)^{\alpha+\beta-1}B(\alpha,\beta)} \mathrm{d}\rho\,.
\end{equation}
Then we have $P(\rho | \Delta_a=x,\Delta_b=y)$ equal to:
\begin{equation}\footnotesize
\left(\dfrac{(b-a+x)^\alpha \twoFone(\alpha,1-\beta,\alpha+1,\frac{b-a+x}{b-a+x+y})- x^\alpha \twoFone(\alpha,1-\beta,\alpha+1,\frac{x}{b-a+x+y})}{\alpha (b-a+x+y)^{\alpha} B(\alpha,\beta)}\right)^k\,.
\end{equation}
The joint $P(\rho, \Delta_a\leq \frac{\delta}{2},\Delta_b\leq \frac{\delta}{2})$ can thus be obtained by the following integral:
\begin{equation}
\footnotesize
\gamma \!\!\int_{0}^{\delta/2}\!\!\!\int_{0}^{\delta/2}\!\!\left(\dfrac{(b-a+x)^\alpha \twoFone(\alpha,1-\beta,\alpha+1,\frac{b-a+x}{b-a+x+y})- x^\alpha \twoFone(\alpha,1-\beta,\alpha+1,\frac{x}{b-a+x+y})}{\alpha (b-a+x+y)^{\alpha} B(\alpha,\beta)}\right)^k
\!\!\!
\mathrm{d}x \mathrm{d}y
\,.
\end{equation}
The marginal distribution for $\rho$ can also be obtained by solving the following integral:
\begin{align*}
P(\rho)= \gamma \int_{0}^{a+(1-b)}\int_{0}^{a+(1-b)-y}P(\rho | \Delta_a=x,\Delta_b=y) \mathrm{d}x \mathrm{d}y\,.
\end{align*}
The l.h.s. of Eq.~\eqref{eq:confidenceBeta} is just the ratio between $P(\rho, \Delta_a\leq \frac{\delta}{2},\Delta_b\leq \frac{\delta}{2})$ and $P(\rho)$.
\end{proof}
\begin{proof}[of Cor.~\ref{cor:newbounds}]
For $\alpha=\beta=1$, $\twoFone(\alpha,1-\beta,\alpha+1,\mu)=1$, $B(\alpha,\beta)=1$ and hence:
\begin{equation}
P(\rho |\Delta_a=x,\Delta_b=y)=
\left(\dfrac{b-a+x- x}{b-a+x+y}\right)^n = \left(\dfrac{b-a}{b-a+x+y}\right)^k\,. 
\end{equation}
Eq.~\eqref{eq:bounds_uniform} is finally obtained by computing the integrals as in \citet{zaffalon2021}.
\end{proof}

\begin{proof}[of Cor.~\ref{cor:boundaequalb}]
If $a^*=b^*$ then we have that $a\rightarrow b$, i.e., $L\rightarrow 0$. Moreover we assume here that $\delta \rightarrow 0$ and $\epsilon \rightarrow 1$. We assume that the distribution of the outputs of the EMCC iterations is uniformly distributed in  $[a^*,b^*]$. $P(a=b |\rho)$ is obtained from Eq.~\eqref{eq:bounds_uniform} in the limit $L\to 0$ and $\epsilon \to 1$ and the thesis follows trivially.
\end{proof}

\begin{proof}[of Thm.~\ref{th:unimodal}]
Let us first define a variable $Y:=h(\bm{X})$ where:
\begin{equation}\label{eq:h}
h(\bm{x}) := \left\{ \begin{array}{ll} \bm{x}&\mathrm{if}\, g(\bm{x})=1\,,\\ *& \mathrm{otherwise}\,.\end{array}\right.
\end{equation}
The states of $Y$ are those in $\Omega_{Y}:= \Omega_{\bm{X}}^{S=1} \cup \{*\}$ with $\Omega_{\bm{X}}^{S=1}:=g^{-1}(S=1)$. As $\bm{X}=f(\bm{U})$, $Y=l(\bm{U})$ with $l := h \circ f$. This allows to obtain from $M$ PSCM $M'$ with the same exogenous variables, and $Y$ as the only endogenous with SE is $l$. Note that $l$ satisfies surjectivity and $M'$ is therefore a  PSCM s.t. $Y$ is a common child of all the nodes associated with the variables in $\bm{U}$. We similarly obtain from $\mathcal{D}_S$ a dataset $\mathcal{D}'$ of $d$ complete observations of $Y$ as follows: for each (missing) observation of $\bm{X}$ in $\mathcal{D}_0$ (whose number is $d_0$) we add an observation $Y=*$ to $\mathcal{D}'$ (remember that $*$ is a proper state of $Y$); while the observations of $\bm{X}$ in $\mathcal{D}_1$ are directly added to $\mathcal{D}'$, where they are regarded as observations of $Y$ instead of $\bm{X}$. Denote the latter set as $\mathcal{D}_1'$. The marginal log-likelihood of $\mathcal{D}'$ for $M'$ is:
\begin{equation}\label{eq:mll3}
LL' := \log P(\mathcal{D}') = \sum_{y\in\mathcal{D}'} P(y) = d_0 \log P(Y=*) + \sum_{\bm{x}\in\mathcal{D}_1'} \log P(Y=\bm{x})\,.
\end{equation}

As $Y=*$ if and only if $S=0$ and, by construction, $\mathcal{D}_1=\mathcal{D}_1'$, writing the log-likelihood of $\mathcal{D}'$ as a function of the exogenous PMFs gives:
\begin{equation}\label{eq:mll4}
LL[\{P(U)\}_{U\in\bm{U}}] = d_0 \log P(S=0) + \sum_{\bm{x}\in\mathcal{D}_1} \log \sum_{\bm{u}\in\Omega_{\bm{U}}} 
\delta_{l(\bm{u}),\bm{x}}
P(\bm{u})\,.
\end{equation}
Yet, $\bm{x}\in\mathcal{D}_1$ means $\bm{x}\in\Omega_{\bm{X}}^{S=1}$ and hence $g(\bm{x})=1$. On these values $h$ in Eq.~\eqref{eq:h} acts as the identical map, and $l(\bm{u})=h \circ f(\bm{u})=f(\bm{u})$. This proves that functions in Eqs.~\eqref{eq:mll4} and \eqref{eq:llu} coincide, and hence that the log-likelihood of $\mathcal{D}_S$ for $M$ coincides with the log-likelihood of $\mathcal{D}'$ for SCM $M'$. Applying Thm.~1 in \citet{zaffalon2021} to such a log-likelihood proves that it has no local maxima and a single global maximum value; that is the first claim.

To prove the equivalence between the M-compatibility constraints and the log-likelihood achieving its (global) maximum value $LL^*$, we start from the direct implication. 
This easily follows by putting Eqs.~\eqref{eq:sm_compat1} and~\eqref{eq:sm_compat2} in Eq.~\eqref{eq:llu} as this gives
Eq.~\eqref{eq:ll_star}. For the converse, assume \emph{ad absurdum} an incompatible specification giving value $LL^*$. The two terms in the r.h.s. of Eq.~\eqref{eq:llu} cannot coincide with those in the r.h.s. of Eq.~\eqref{eq:ll_star} otherwise the model would be compatible. One of them should be greater than the corresponding term in $LL^*$: but this is impossible as both terms of $LL^*$ have been obtained by independent maximisation of the log-likelihood functions associated with $\mathcal{D}_0$ resp. $\mathcal{D}_1$. This proves the equivalence between the compatibility constraints and the fact that the log-likelihood is equal to $LL^*$. It also proves that $LL^*$ is actually the global maximum of the log-likelihood, since incompatible models return values smaller than $LL^*$, while compatible ones match $LL^*$ exactly.
\end{proof}

\acks{
\indent This research was partially funded by MCIN/AEI/10.13039/501100011033 with FEDER funds for the project PID2019-106758GB-C32, and also by Junta de Andaluc\'{i}a grant P20-00091. Finally we would like to thank the ``Mar\'{i}a Zambrano'' grant (RR\_C\_2021\_01) from the Spanish Ministry of Universities and funded with NextGenerationEU funds.}


\end{document}